\UseRawInputEncoding
\pdfoutput=1
\documentclass[10pt,twocolumn,twoside]{IEEEtran}
\usepackage{multirow}
\usepackage{bm}
\usepackage[cmex10]{amsmath}
\usepackage{amssymb}
\usepackage{amsthm}
\usepackage{graphicx}
\usepackage{subfigure}
\usepackage{multirow}
\usepackage[hyphens]{url}

\theoremstyle{remark}
\newtheorem{remark}{Remark}

\usepackage{algorithm}
\usepackage{algorithmicx}
\usepackage{algpseudocode}
\usepackage{verbatim}
\usepackage{booktabs}
\usepackage{threeparttable}
 
\usepackage{cite}
\usepackage[numbers,sort&compress]{natbib}
\usepackage{overpic}
\usepackage{rotating}
\usepackage{amsmath}

\begin{document}

	\title{On the Connection of Generative Models and Discriminative Models for Anomaly Detection}

	\author{ Jingxuan~Pang and Chunguang~Li,~\IEEEmembership{Senior~Member,~IEEE}
		\thanks{The authors are with the College of
			Information Science and Electronic Engineering, Zhejiang
			University, Hangzhou 310027, China, and also with the Ningbo Research Institute, Zhejiang University, Ningbo 315100, China  (C. Li is
			the corresponding author, email: cgli@zju.edu.cn).} }
	\maketitle
	
	\begin{abstract}
		Anomaly detection (AD) has attracted considerable attention in both academia and industry. Due to the lack of anomalous data in many practical cases, AD is usually solved by first modeling the normal data pattern and then determining if data fit this model. Generative models (GMs) seem a natural tool to achieve this purpose, which learn the normal data distribution and estimate it using a probability density function (PDF). However, some works have observed the ideal performance of such GM-based AD methods. 
		In this paper, we propose a new perspective on the ideal performance of GM-based AD methods. We state that in these methods, the implicit assumption that connects GMs'results to AD's goal is usually implausible due to normal data's multi-peaked distribution characteristic, which is quite common in practical cases. We first qualitatively formulate this perspective, and then focus on the Gaussian mixture model (GMM) to intuitively illustrate the perspective, which is a typical GM and has the natural property to approximate multi-peaked distributions.
		Based on the proposed perspective, in order to bypass the implicit assumption in the GMM-based AD method, we suggest integrating the Discriminative idea to orient GMM to AD tasks (DiGMM). 
		With DiGMM, we establish a connection of generative and discriminative models, which are two key paradigms for AD and are usually treated separately before. This connection provides a possible direction for future works to jointly consider the two paradigms and incorporate their complementary characteristics for AD.
	\end{abstract}

	
	\maketitle	
	
	\vspace{3mm}
	\section{Introduction} \label{sec:1}
	\IEEEPARstart
	Anomaly detection aims at discovering data that deviate from expected behavior, which has been a hot issue in both academia and industry for its important role in ensuring system reliability and stability \cite{Chandola2009, miao2019, liu2020, zhou2021, pang2022}.
	As the systems are usually complex in practical anomaly detection tasks, data-driven anomaly detection has gained considerable attention thanks to the dramatic increase of data \cite{Ramaswamy2000,kurt2022,Scholkopt2001}. Such methods do not rely on a priori knowledge of the system model, but utilize historically recorded data to construct anomaly detectors. As it is usually easy to obtain normal data but hard to obtain anomalies in most practical scenarios, a popular practice is to use the recorded normal data to mine the normal data pattern and identify data that does not fit this pattern as anomalies \cite{Scholkopt2001,Tax1999,li2021,wang2018,kong2021}. 
	Generative models are natural to achieve this purpose and have motivated many GM-based AD methods, as they aim to model the normal data pattern by estimating data distribution \cite{wang2018,kong2021,Pidhorskyi2018}. These methods often first learn a GM to describe the normal data distribution using a probability density function (PDF), with which the probability density of data can be estimated. Then, a threshold is set for the PDF value to establish the anomaly detector. Such a detector will recognize data points with lower PDF values than the set threshold as anomalies.
	
	However, several works have observed that GM-based methods do not show ideal performance in AD tasks \cite{Nalisnick2019, Hendrycks2019, Choi2018}. 
	Especially, they observe that many GMs assign higher estimated probability densities to anomalies than normal data, thereby causing misidentification. This observation seemingly indicates the infeasibility of directly applying GMs to AD. 
	There have been a few works attempting to explain this phenomenon, and most of them focus mainly on deep GMs \cite{Nalisnick2019-2, Wang2020, Zhang2021, Kirichenko2020, Le2021}. 
	Specifically, the authors in \cite{Nalisnick2019-2} attribute the ideal performance to a mismatch between the model¡¯s typical set \cite{typicalset} and the areas of high probability density. They state that anomaly detection should be implemented by checking if data resides in the former, while previous methods often check if data falls in the latter.
	The authors in \cite{Wang2020} also explain the ideal performance in terms of the typical set, and propose a new outlier test that generalizes the idea of the typical set test.
	The authors in \cite{Zhang2021} understand the anomaly detection failures of deep GMs as the model estimation error. They state that some small-volume regions of the sample space are unimportant in generative models but should be accurately estimated in anomaly detection. Due to the poor estimation over these regions, good generation is not sufficient for good anomaly detection.
	In terms of the specific method, the authors in \cite{Kirichenko2020} focus mainly on the reason why normalizing flows (NFs) \cite{Tabak2013,Dinh2015,Dinh2017} fail for anomaly detection. They show that it is difficult for NFs to detect data with anomalous semantics, since they learn latent representations of data based on local pixel correlations instead of semantic content.
	Besides, the authors in \cite{Le2021} show that distribution densities carry less meaningful information for anomaly detection than previously thought. They point out that it relies on strong and implicit hypotheses to use the probability density given by GMs for anomaly detection, and they emphasize the necessity of incorporating prior knowledge into GMs.

	In this article, we propose a new perspective on the ideal performance of GM-based AD methods. These methods usually assume the probability density as an indicator of the possibility that a sample is normal \cite{wang2018,kong2021,Pidhorskyi2018}, however, we state that this assumption often does not hold due to the multi-peaked characteristic of the normal data distribution, which is very common in practical scenarios. Specifically, the probability density is not exactly positively correlated with the possibility that a sample is normal, thereby cannot always act as an indicator of the latter.
	For this reason, GMs, which aim at learning the underlying probability density of normal data and approximating it by a PDF, are not directly suitable for anomaly detection by setting a threshold for the PDF value. That is, even if a GM is well-established to fit the normal data distribution, using it directly for anomaly detection may not achieve good performance.
	To summarize the proposed perspective, the implicit assumption connecting  GMs' results to AD's goal is not valid, thereby causing the ideal performance of GMs in AD.
	We first qualitatively formulate this perspective, and then focus on the Gaussian mixture model (GMM) to intuitively illustrate the perspective, which is a typical GM and has the natural property to approximate multi-peaked distributions.

	Based on the proposed perspective, to bypass the implicit assumption in GMM-based AD methods, we orient GMM towards AD tasks by integrating the discriminative objective, i.e., discrimination-integrated GMM (DiGMM). This is inspired by the task-oriented property of discriminative learning that it directly learns a decision boundary for anomaly detection guided by the detecting goal \cite{Juang1992,Brown2010,Del2016}.
	We leverage this property to learn the decision boundary based on GMM¡¯s result.  Specifically, we first specify the decision boundary based on the discriminative formulation of GMM. Then, we formulate an optimization problem using the discriminative goal to learn the decision boundary. In this way, the learned decision boundary leverages both the normal data patterns extracted by GMM and the characteristics of discriminative learning. Moreover, with DiGMM, we develop a joint perspective on GMM and the one-class support vector machine (OCSVM), where the latter is a typical discriminative model for AD. Thus, we offer a way to connect generative models and discriminative models for AD, which are usually treated as two separate paradigms before.

	The rest of this paper is organized as follows. In Section \uppercase\expandafter{\romannumeral2}, some necessary preliminary knowledge is briefly introduced to make the paper self-contained. Afterward, in \uppercase\expandafter{\romannumeral3}, we qualitatively interpret the proposed perspective and focus on GMM to intuitively illustrate it. Then, in Section \uppercase\expandafter{\romannumeral4}, we propose a method named DiGMM and describe the connection of GMM and OCSVM. Finally, discussions are given in Section \uppercase\expandafter{\romannumeral5}.
	
	\emph{Notations:}
	In this paper, we use normal letters, lowercase boldface, and uppercase boldface to denote scalars, vectors, and matrices, respectively. Besides, $(\cdot)^T$ denotes transposition, and $\|\cdot\|$ stand for the $L_2$-norm. Other notations will be introduced when necessary.

	\vspace{3mm}
	\section{Preliminary}	
	\subsection{Gaussian Mixture Model}
	
	In practical scenarios, the distribution of normal data usually is complex and diversifies in different datasets. Moreover, the distribution is usually unknown in advance, so a flexible model is needed to model the normal data distribution. Gaussian Mixture Model (GMM) can flexibly model data distributions using a combination of several Gaussian distributions, making it a classic and commonly used probability-based generative method for modeling the data distribution \cite{Reynolds2000}. In this section, we will introduce GMM in detail.
	
	GMM models the whole distribution of data $\mathcal{X}$ using a mixture of Gaussian distribution. Each distribution is parameterized by the mean and the covariance matrix, and it is commonly called as a component. Specifically, GMM describes the data distribution using the probability density function (PDF)
	\begin{equation}
	p(\bm{x}|\Theta)=\sum_{j=1}^{m} w_{j} p_j(\bm{x}|\bm{\mu}_j,\bm{\Sigma}_j),
	\end{equation}
	where $\bm{x}$ denotes a certain sample and $\Theta$ collectively represents all parameters as $\Theta = \{ w_j,\bm{\mu}_j,\bm{\Sigma}_j , j = 1,2,...,m\}$.
	At the right end of the equation, $w_j$ denotes the mixing weight of the $j$th Gaussian components over all $m$ components and $\sum_{j=1}^{m} w_{j}=1$,
	and $p_j(\bm{x}|\bm{\mu}_j,\bm{\Sigma}_j)$ represents the PDF of the $j$th Gaussian component $\mathcal{N}(\bm{\mu}_j,\bm{\Sigma}_j)$, and its form is
	\begin{equation}\label{eq5}
	p_j(\bm{x}|\bm{\mu}_j,\bm{\Sigma}_j) = \frac{\exp(-\frac{1}{2}(\bm{x} - \bm{\mu}_j)^T\bm{\Sigma}_j^{-1}(\bm{x} - \bm{\mu}_j))}{(2\pi)^{\frac{d}{2}}(\det(\bm{\Sigma}_j))^{\frac{1}{2}}},
	\end{equation}
	where $\bm{\mu}_j$ and $\bm{\Sigma}_j$ denote the mean and covariance matrix of the $j$th Gaussian component, respectively, and $d$ denote the data dimension.
	
	So far the parametric form and parameters of GMM's PDF have been determined, and then the goal is to estimate the value of these parameters so that the specific form of GMM's PDF can be formulated for a certain data set. It is typical to adopt Maximum Likelihood Estimation (MLE), which estimates model parameters using available data.
	
	Denote the training normal data set as $\mathcal{X}=\{\bm{x}_i,i=1,2,\ldots,n\}$, where $\bm{x}_i\in \mathbf{R}^d$. MLE estimates $\Theta$ so that $p(\mathcal{X})$ is maximum. By introducing the log-likelihood function
	\begin{equation}
	\begin{split}
	\mathcal{L} & = {\rm log} \, p(\mathcal{X}|\Theta) = {\rm log} \prod_{i=1}^{n} p(\bm{x}_i|\Theta) \\
	&= \sum_{i=1}^{n} {\rm log} \left(   \sum_{j=1}^{m} w_{j} p_j(\bm{x}_i|\bm{\mu}_j,\bm{\Sigma}_j) \right),
	\end{split}
	\end{equation}
	the problem can be mathematically described as $\Theta = {\rm arg \, max}_{\Theta} \mathcal{L}$. Expectation Maximization (EM) is the most commonly used technique to solve this problem and find an approximated solution through iteration.

	\begin{figure}[h]
		\centering
		\includegraphics[scale=0.5]{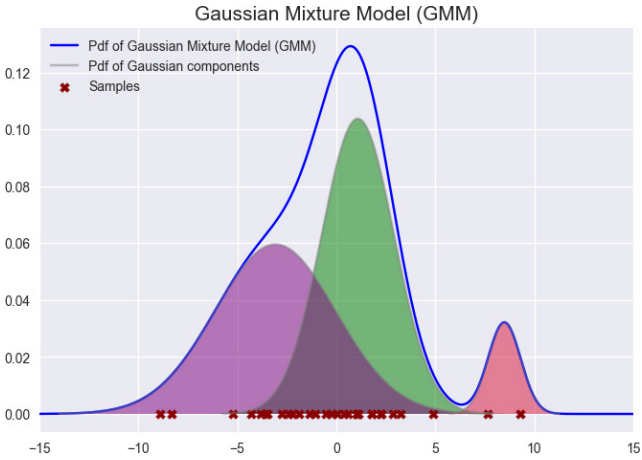}
		\caption{$\left[\text{color online}\right]$ Illustration of Gaussian Mixture Model (GMM).}
	\end{figure}

	\vspace{3mm}
	\section{A new perspective on the ideal performance of GMs in AD} 
	In this section, we propose a new perspective to provide a possible explanation for the unsatisfactory performance of GM-based AD methods. These methods usually implicitly assume the probability density as an indicator of the possibility that a sample is normal, thereby recognizing data as anomalies if their estimated probability densities are lower than a certain value. In this section, we will present the implausibility of this assumption in terms of the multi-peaked distribution characteristic of normal data, which is quite common in practical scenarios. 
	
	First, consider an ideal case that normal data are distributed simply as a single-peaked distribution such as Gaussian distribution. Intuitively, given a new data point, the higher its probability density, the higher the possibility that it is normal. The implicit assumption holds in such an ideal case. 
	However, in practical scenarios, the underlying normal data distribution is usually complex, and maybe contains irregularly changing density, multiple classes, and so on. It rarely satisfies the condition of the single-peaked distribution, i.e., the data density monotonically decreases as data get farther to a certain sample along any direction. That is, the normal data distribution is usually a multi-peaked distribution. Besides, the peak values are usually varying.  In such cases, a normal sample around one peak may have the same or smaller probability density than an anomaly around another peak, since the normal data around different peaks have different ranges of probability densities. That means that the probability density is not exactly positively correlated with the possibility that a sample is normal, thereby cannot always act as an indicator of the latter.
	Based on the above analysis, GMs, which aim at estimating the data probability density, are not directly applicable for anomaly detection by additionally setting a threshold for the estimated probability density (i.e., PDF values). To summarize the proposed perspective, the implicit assumption connecting  GMs' results to AD's goal is not valid due to the multi-peaked data distribution, thereby causing the ideal performance of GMs in AD.
	
	\begin{figure*}[h]
		\centering
		\subfigure[]{
			\label{gmm:subfig:a}
			\includegraphics[width=5.8cm]{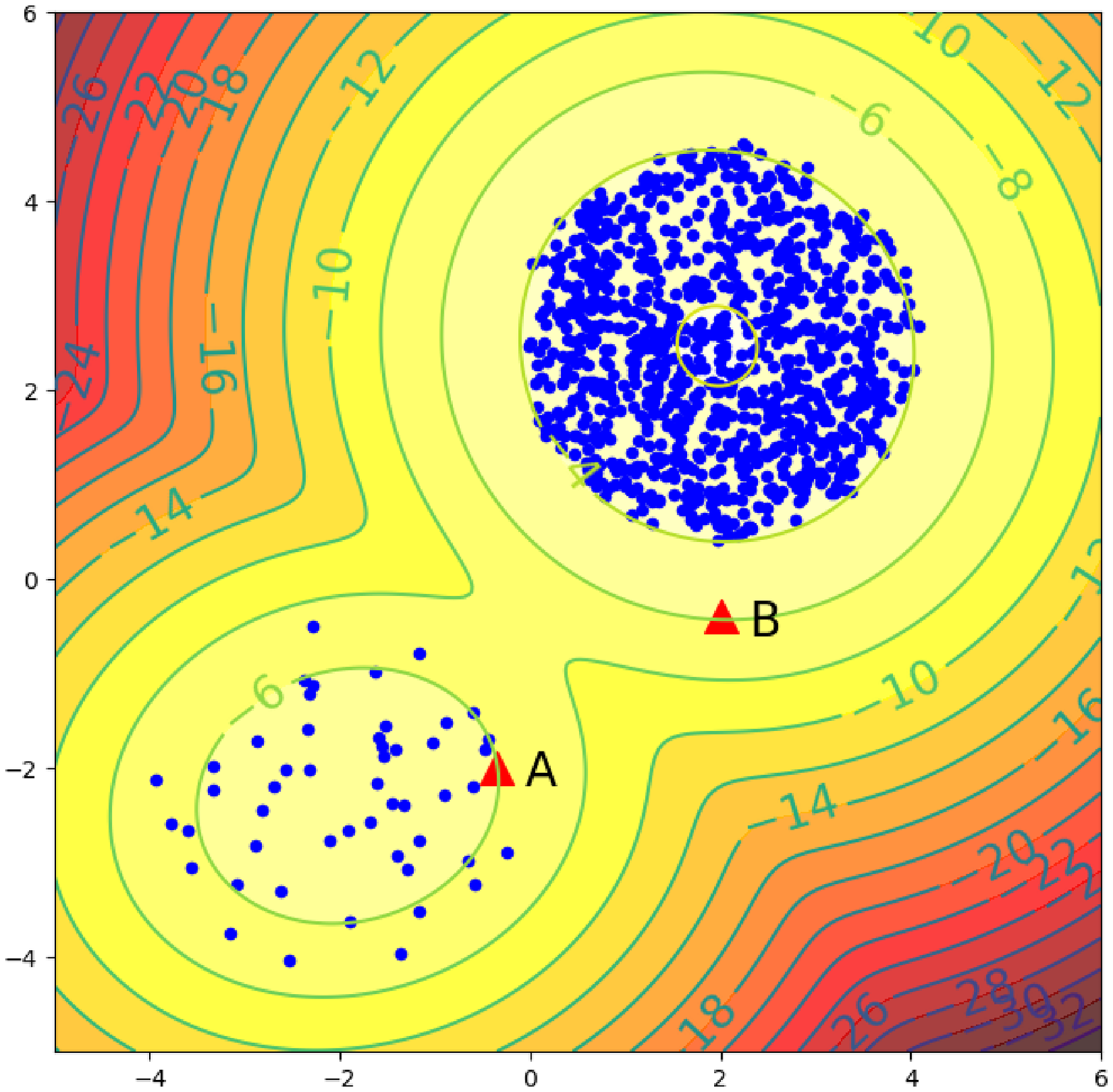}}
		\subfigure[]{
			\label{gmm:subfig:b}
			\includegraphics[width=5.8cm]{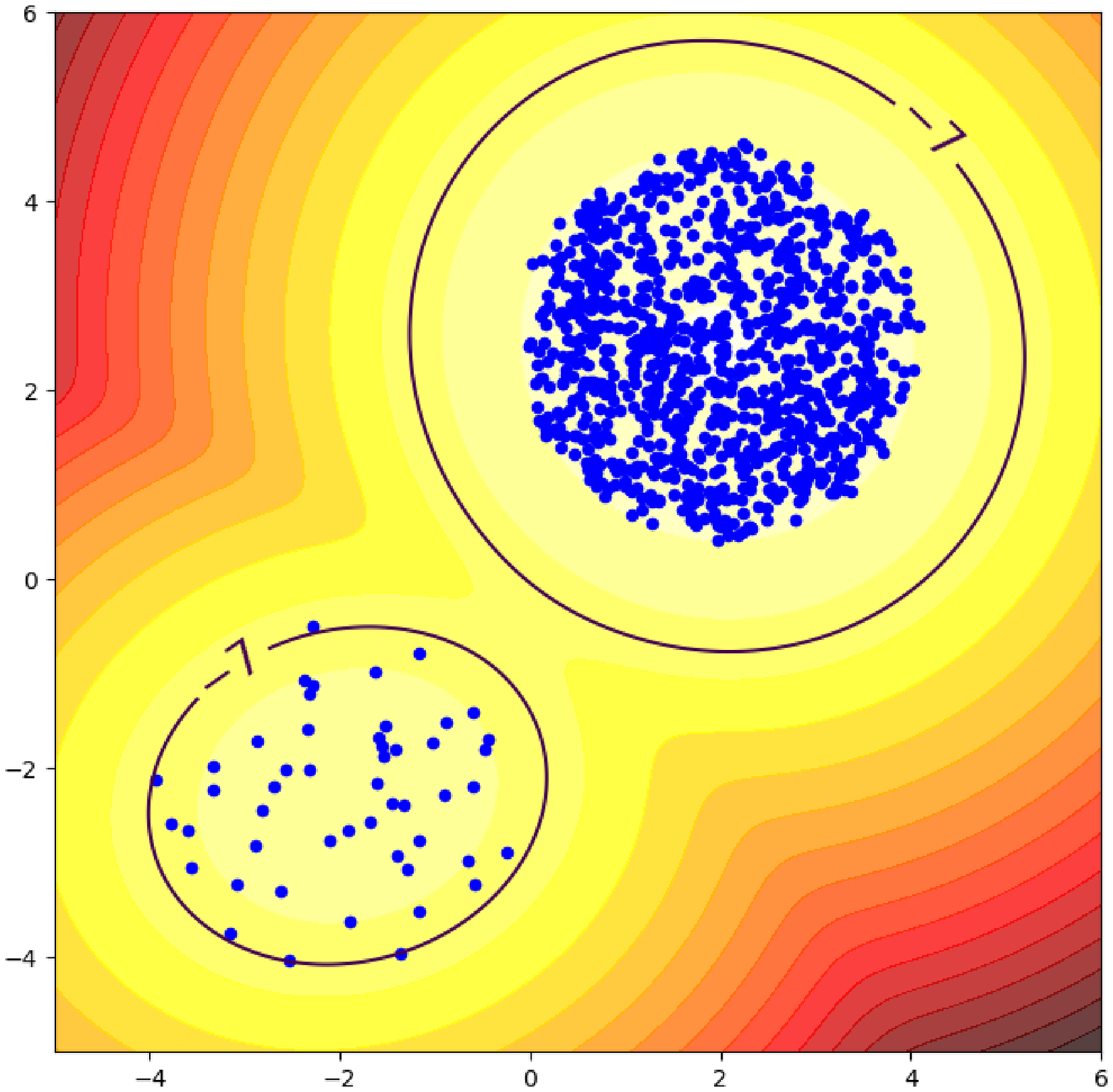}}
		\subfigure[]{
			\label{gmm:subfig:c}
			\includegraphics[width=5.8cm]{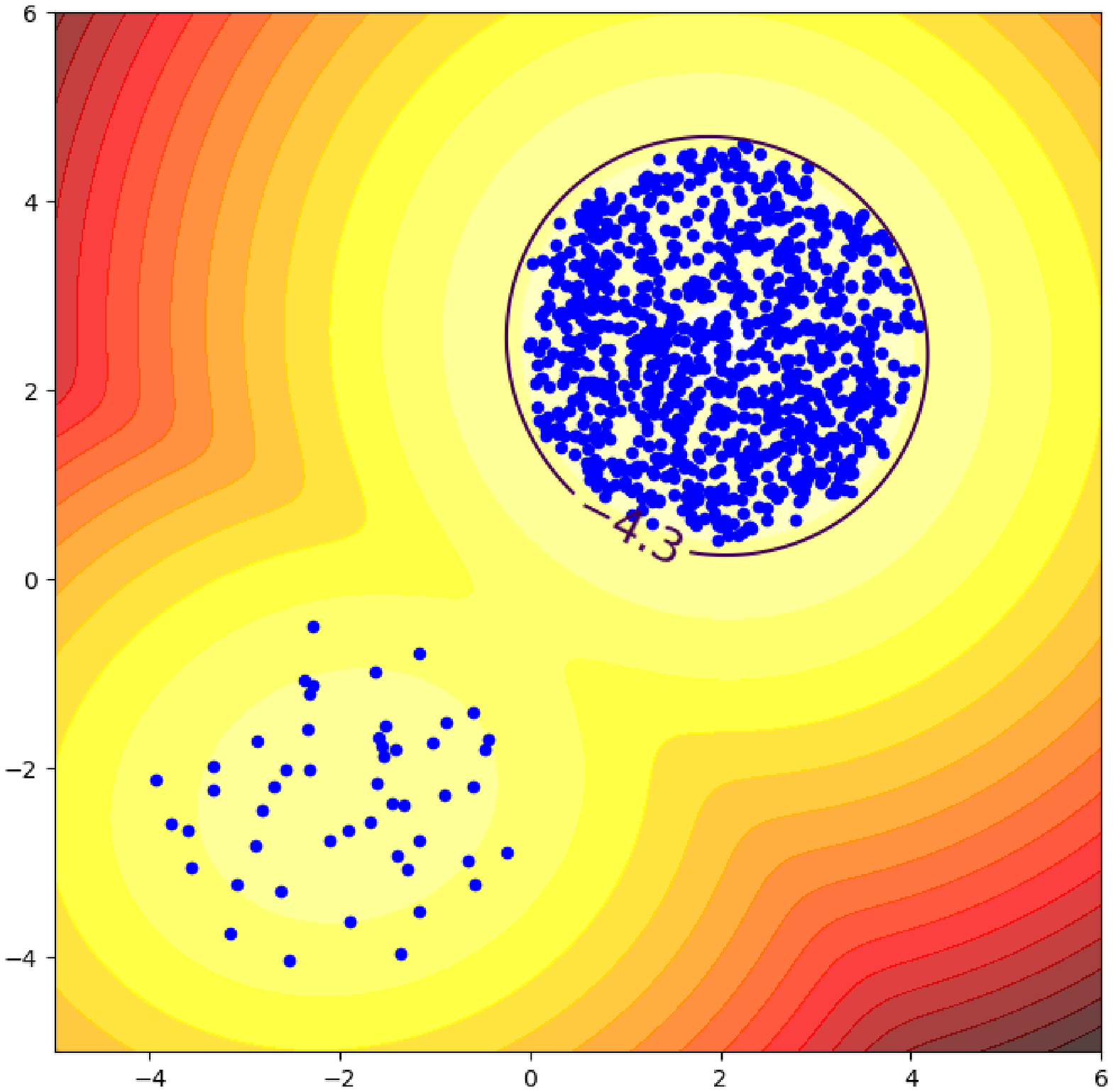}}
		\vspace*{-1mm}
		\caption{$\left[\text{color online}\right]$ Results of GMM. The blue solid circles represent normal training samples, and the curves with value labels represent the contour of the natural logarithm of the PDF value given by GMM. The two red triangles in the subfigure (a) represent two test data points lying on the same contour curve.}
		\label{gmm}	
	\end{figure*}
	
	We now focus on GMM, a typical and widely used GM, to illustrate the above perspective. For visualization, we generate two-dimensional synthetic data and treat them as normal samples to build GMM models. The data and GMM's results are shown in Fig. \ref{gmm:subfig:a}, where the blue solid circles represent generated normal samples, and the green curves with value labels represent the contour of the natural logarithm of the PDF value given by GMM. The red triangles represent two test data points A and B. It is shown that point A tends to be a normal sample since it is very close to the known normal samples, while point B tends to be an anomalous one. However, these two points have the same PDF value $e^{-6}$, which shows the inconsistency of the size of the PDF value with the possibility of a sample being normal. 
	Then, we consider the anomaly detector constructed by directly setting a threshold for the PDF value given by GMM.
	For intuition, we use a curve to visualize the decision boundary defined by the threshold of the PDF value, and observe how well the boundary encloses the normal samples. Note that a good boundary for anomaly detection should enclose as many as the training normal data tightly.
	Due to the inconsistency of the size of the PDF value with the possibility of a sample being normal, it is difficult for the detector to find a good boundary. 
	For instance, if the threshold is set to be  $e^{-7}$, as shown in Fig. \ref{gmm:subfig:b}, both normal and anomalous data near the left sparse region can be well recognized, but lots of anomalies near the dense region will be wrongly recognized as normal data. As the set threshold gradually increases, samples in dense regions are gradually better identified, but more normal samples in sparse regions will be identified as abnormal. If the threshold is set to be $e^{-4.3}$, as shown in Fig. \ref{gmm:subfig:c}, although data near the sparse region can be well recognized, all normal data in the sparse region will be wrongly recognized as anomalies, since the PDF values of data in the sparse region are all relatively smaller. That is, it is difficult to obtain a good decision boundary in all regions by setting a threshold for the PDF value given by GMM.

	\vspace{3mm}
	\section{From GMM to OCSVM }
	Based on the perspective in Section \uppercase\expandafter{\romannumeral3}, in this section, we will advise an AD method integrating Discrimination and GMM (DiGMM), with which we establish a connection of GMM and OCSVM. 
	
	Recalling the form of the discriminative rule of GMM for AD, it comprises the learned PDF value and a globally uniform threshold. As stated in section \uppercase\expandafter{\romannumeral3}, such a rule is ineffective for AD. In order to learn an effective discriminative rule for AD, we integrate discriminative ideas into the learning of the decision boundary, which is inspired by discriminative learning's task-oriented property of focusing on finding a decision boundary to construct the discriminative rule. The method is introduced in detail in the following.
	
	\vspace{2mm}
	\textbf{Problem Formulation.} We first specify the discriminative formulation of AD based on the results of GMM. Once the GMM model is established, the normal data distribution is described by the PDF formulated as $p(\bm{x})=\bm{w}^T\bm{p}(\bm{x})$, where $\bm{p}(\bm{x})$ denotes the vector assembling PDFs of all Gaussian components $ \mathcal{N}(\bm{\mu}_j,\bm{\Sigma}_j), j=1,2,...m$, that is, 
	
	\begin{equation}
	p(\bm{x})=\bm{w}^T\bm{p}(\bm{x})=\sum_{j=1}^{m} w_{j} p_j(\bm{x}|\bm{\mu}_j,\bm{\Sigma}_j).
	\end{equation}
	Denote the threshold for the PDF value as $\rho$, then the decision boundary that separates normal data from anomalous data is 
	\begin{equation}\label{decision boundary}
	\bm{w}^T\bm{p}(\bm{x})-\rho=0.
	\end{equation}
	The corresponding decision function is formulated as $f(\bm{x})=\bm{w}^T\bm{p}(\bm{x})-\rho$. That is, given a data point $\bm{x}$, it will be labeled as a normal sample if $f(\bm{x})>0$, and an abnormal one otherwise.
	
	Now we consider adjusting the decision boundary  \eqref{decision boundary} to overcome the problem stated in section \uppercase\expandafter{\romannumeral2} and improve AD performance. Recall that the overall PDF $p(\bm{x})$ of GMM cannot reflect the normal possibility due to the multi-peaked characteristic of the normal data distribution. However, each Gaussian component of GMM is single-peaked, and its PDF value can to some extent reflect the normal possibility in terms of this component. Therefore, in order to leverage this helpful information, we keep the Gaussian components $\bm{p}(\bm{x})$ unchanged in \eqref{decision boundary}, and jointly learn parameters $\bm{w}$ and $\rho$ guided by discriminative ideas.
	
	Then, our goal is to construct the optimization problem to learn parameters $\bm{w}$ and $\rho$. Based on discriminative ideas, the aim of the optimization problem is to make the decision boundary recognize as more as the training normal data to be normal and fall as close as possible to these data. For this purpose, we will formulate the optimization problem considering the following aspects.
	
	1) Considering the discriminative results $\bm{w}^T\bm{p}(\bm{x})-\rho$ should be positive for as many training normal samples as possible, we impose penalties to the training data that give negative discriminative results by the slack penalty term $max\{\rho-\bm{w}^T\bm{p}(\bm{x}), 0\}$.
	
	2) In order to push the decision boundary as close as possible towards the training normal points, we subtract $\rho$ to make it bigger.
	
	3) Moreover, we add the regularization term $\frac{1}{2}\|\bm{w}\|^2$ into the objective function.
	
	To sum up, the overall objective function is formulated as follows.
	\begin{equation}\label{optimizaiton problem}
	\mathop{\min}_{\bm{w},\rho} \quad  \frac{1}{2}\|\bm{w}\|^2 +  \frac{1}{\nu n}\sum_{i=1}^{n} {\rm max} \{(\rho - \bm{w}^T  \bm{p}(\bm{x}_i)),0\}-\rho,
	\end{equation} 
	where $n$ denotes the number of all training normal data, and $\nu$ is the penalty hyperparameter.
	
	\vspace{2mm}
	\textbf{From GMM to OCSVM.} The optimization problem \eqref{optimizaiton problem} precisely has the same form as that of the one-class support vector machine (OCSVM) \cite{Scholkopt2001}. In both \eqref{optimizaiton problem} and the optimization problem of OCSVM, $\bm{w}$ and $\rho$ are parameters that need to be learned. $\bm{p}(\bm{x})$ in \eqref{optimizaiton problem} denotes the assembled vector of PDFs of all Gaussian components, and it corresponds to the mapping function in OCSVM that maps the input samples into the reproducing kernel Hilbert space. We elaborate that $\bm{p}(\bm{x})$ in \eqref{optimizaiton problem} is able to act as the mapping in OCSVM in \emph{Remark} \ref{remark}. Thus, we establish a connection of GMM and OCSVM for AD not only in form but also in theory.
	
	\begin{remark} \label{remark}
		For the vector $\bm{p}(\bm{x})$ in \eqref{optimizaiton problem}, each of its dimensions describes the possibility that a sample is generated by a certain Gaussian component. Intuitively, for a normal sample, there is at least one distribution that has a high possibility of generating this sample, making it far from the origin in the feature space. On the contrary, for an anomaly, since the possibility it comes from any distribution is relatively small, it is close to the origin in the feature space. In addition, as the values of all dimensions are non-negative, all pairwise angles between samples in the feature space are in the range of 0$^{\circ}$ to 90$^{\circ}$. The above two properties enable the $\bm{p}(\bm{x})$ in \eqref{optimizaiton problem} to be regarded as the mapping in OCSVM.
	\end{remark}
	
	\vspace{2mm}
	\textbf{Problem Solution.} Based on the above connection, \eqref{optimizaiton problem} can be solved referring to OCSVM \cite{Scholkopt2001} to obtain the optimal parameters $\bm{w}^*$ and $\rho^*$. Then, the decision boundary is constructed as $\bm{w}^{*T} \bm{p(x)}-\rho^*=0$, which separates normal data from anomalous ones. The complete procedure is summarized in Algorithm 1.
	\begin{algorithm}[h]
		\caption{DiGMM for Anomaly Detection}
		\begin{enumerate}
			
			\item[a)] \textbf{Training Phase:}
			
			\textbf{Input:} Normal sample set $\mathcal{X}$ and the hyperparameter $\nu$.
			\begin{enumerate}	
				\item[1)] Implement GMM on $\mathcal{X}$ to calculate parameters of all Gaussian components, i.e., the mean vectors $\bm{\mu}_j$ and the covariance matrices $\bm{\Sigma}_j$.
				\item[2)] Construct vectors $\bm{p}(\bm{x})$ for all data by assembling PDF values of all Gaussian components $ \mathcal{N}(\bm{\mu}_j,\bm{\Sigma}_j), j=1,2,...m$.
				\item[3)] Solve the optimization problem \ref{optimizaiton problem} referring to OCSVM to calculate the optimal $\bm{w}^*$ and $\rho^*$.
				\item[4)] Construct the decision function $f(\bm{x}) = \bm{w}^{*T} \bm{p}(\bm{x})-\rho^*$.
			\end{enumerate}
			
			\item[b)] \textbf{Detecting Phase:}
			
			\textbf{Input:} Sample set $\mathcal{X}$ that need to be detected, the mean vectors $\bm{\mu}_j$ and the covariance matrices $\bm{\Sigma}_j$ given by GMM, the the decision function $f(\bm{x}) = \bm{w}^{*T} \bm{p}(\bm{x})-\rho^*$.
			
			\begin{enumerate}
				\item[1)] Calculate $\bm{p}(\bm{x})$ for each sample $\bm{x}$ in $\mathcal{X}$ by calculating and assembling PDF values of all Gaussian components.
				\item[2)] Compute the decision function for each sample, and the sample $\bm{x}$ will be labeled as a normal sample if $f(\bm{x})>0$, and an abnormal one otherwise.	
			\end{enumerate}
		\end{enumerate}
	\end{algorithm}

	\begin{figure}[h]
		\centering
		\includegraphics[scale=0.4]{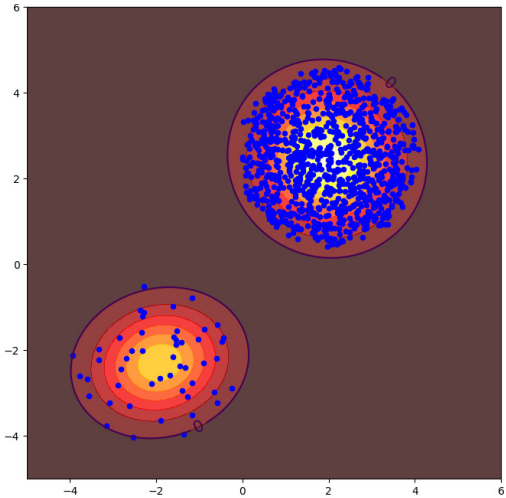}
		\caption{$\left[\text{color online}\right]$ Results of DiGMM. Blue solid circles represent normal training data, and the curve with label 0 denotes the decision boundary obtained by DiGMM.}
		\label{digmm}
	\end{figure}
	
	\vspace{2mm}
	\textbf{Analysis.} The proposed DiGMM leverages both the normal data patterns extracted by GMM and the task-oriented property of discriminative learning. 
	Benefiting from the latter, DiGMM bypasses the implausible assumption stated in Section \uppercase\expandafter{\romannumeral3}, and has the potential to correctly identified data around different peaks of multi-peaked distribution. 
	To visually verify the effectiveness of the proposed method for AD, we apply the proposed method on the two-dimensional synthetic data in Section \uppercase\expandafter{\romannumeral3}, and the results are shown in Figure \ref{digmm}. The curve with label 0 denotes the decision boundary obtained by DiGMM. As shown in the figure, the decision boundary encloses normal data in both dense and sparse regions appropriately, which means the decision function is able to correctly discriminate whether newly-given samples are anomalies or not. This indicates that the proposed anomaly detector overcomes the problem shown in Figure \ref{gmm}.

	
	\vspace{3mm}
	\section{Discussion}
	In this paper, we proposed a new perspective on the ideal performance of GMs in AD, and provide a new insight on the connection of generative and discriminative models for AD.
	
	We stated that in GM-based AD methods, the implicit assumption that connects GMs' results to AD's goal usually does not hold due to normal data's multi-peaked distribution characteristic, which is quite common in practical scenarios. Specifically, the probability density that GMs aim to approximate is not exactly positively correlated with the normal possibility of a sample. Thus, if directly using the probability density estimated by a GM and then setting a threshold to construct an anomaly detector, the detector is likely to misidentify samples.
	We qualitatively formulated the proposed perspective, and then focused on GMM to intuitively illustrate the perspective.
	Afterward, we proposed to integrate discriminative ideas with GMM (DiGMM) to bypass the assumption in GMM-based AD methods.
	The integrated discriminative idea enabled us to establish a connection of GMM and OCSVM. Specifically, the discriminative goal allowed us to link the form of GMM to OCSVM, and the intrinsic characteristics of GMM made it converge with OCSVM in theory. 
	Such an insight provides a future direction of connecting generative and discriminative models and jointly considering them for AD, which are usually treated and utilized separately before. We hope that the joint consideration of these two paradigms will enhance anomaly detection by incorporating their complementary properties.

\end{document}